\PassOptionsToPackage{unicode}{hyperref}
\PassOptionsToPackage{hyphens}{url}
\PassOptionsToPackage{dvipsnames,svgnames,x11names}{xcolor}
\documentclass[
]{article}
\usepackage{xcolor}
\usepackage{amsmath,amssymb}
\usepackage{iftex}
\ifPDFTeX
  \usepackage[T1]{fontenc}
  \usepackage[utf8]{inputenc}
  \usepackage{textcomp} 
\else 
  \usepackage{unicode-math} 
  \defaultfontfeatures{Scale=MatchLowercase}
  \defaultfontfeatures[\rmfamily]{Ligatures=TeX,Scale=1}
\fi
\usepackage{lmodern}
\ifPDFTeX\else
\fi
\IfFileExists{upquote.sty}{\usepackage{upquote}}{}
\IfFileExists{microtype.sty}{
  \usepackage[]{microtype}
  \UseMicrotypeSet[protrusion]{basicmath} 
}{}
\makeatletter
\@ifundefined{KOMAClassName}{
  \IfFileExists{parskip.sty}{%
    \usepackage{parskip}
  }{
    \setlength{\parindent}{0pt}
    \setlength{\parskip}{6pt plus 2pt minus 1pt}}
}{
  \KOMAoptions{parskip=half}}
\makeatother
\makeatletter
\ifx\paragraph\undefined\else
  \let\oldparagraph\paragraph
  \renewcommand{\paragraph}{
    \@ifstar
      \xxxParagraphStar
      \xxxParagraphNoStar
  }
  \newcommand{\xxxParagraphStar}[1]{\oldparagraph*{#1}\mbox{}}
  \newcommand{\xxxParagraphNoStar}[1]{\oldparagraph{#1}\mbox{}}
\fi
\ifx\subparagraph\undefined\else
  \let\oldsubparagraph\subparagraph
  \renewcommand{\subparagraph}{
    \@ifstar
      \xxxSubParagraphStar
      \xxxSubParagraphNoStar
  }
  \newcommand{\xxxSubParagraphStar}[1]{\oldsubparagraph*{#1}\mbox{}}
  \newcommand{\xxxSubParagraphNoStar}[1]{\oldsubparagraph{#1}\mbox{}}
\fi
\makeatother

\usepackage{longtable,booktabs,array}
\usepackage{calc} 
\usepackage{etoolbox}
\makeatletter
\patchcmd\longtable{\par}{\if@noskipsec\mbox{}\fi\par}{}{}
\makeatother
\IfFileExists{footnotehyper.sty}{\usepackage{footnotehyper}}{\usepackage{footnote}}
\makesavenoteenv{longtable}
\usepackage{graphicx}
\makeatletter
\newsavebox\pandoc@box
\newcommand*\pandocbounded[1]{
  \sbox\pandoc@box{#1}%
  \Gscale@div\@tempa{\textheight}{\dimexpr\ht\pandoc@box+\dp\pandoc@box\relax}%
  \Gscale@div\@tempb{\linewidth}{\wd\pandoc@box}%
  \ifdim\@tempb\p@<\@tempa\p@\let\@tempa\@tempb\fi
  \ifdim\@tempa\p@<\p@\scalebox{\@tempa}{\usebox\pandoc@box}%
  \else\usebox{\pandoc@box}%
  \fi%
}
\def\fps@figure{htbp}
\makeatother

\NewDocumentCommand\citeproctext{}{}
\NewDocumentCommand\citeproc{mm}{%
  \begingroup\def\citeproctext{#2}\cite{#1}\endgroup}
\makeatletter
 \let\@cite@ofmt\@firstofone
 \def\@biblabel#1{}
 \def\@cite#1#2{{#1\if@tempswa , #2\fi}}
\makeatother
\newlength{\cslhangindent}
\newlength{\csllabelwidth}
\newenvironment{CSLReferences}[2] 
 {\begin{list}{}{%
  \setlength{\itemindent}{0pt}
  \setlength{\leftmargin}{0pt}
  \setlength{\parsep}{0pt}
  \ifodd #1
   \setlength{\leftmargin}{\cslhangindent}
   \setlength{\itemindent}{-1\cslhangindent}
  \fi
  \setlength{\itemsep}{#2\baselineskip}}}
 {\end{list}}
\usepackage{calc}

\providecommand{\tightlist}{%
  \setlength{\itemsep}{0pt}\setlength{\parskip}{0pt}}

\usepackage{fontspec}
\usepackage{polyglossia}

\usepackage[a4paper, margin=1in]{geometry}

\usepackage{xcolor}
\definecolor{arxivblue}{HTML}{000099}
\definecolor{darkslate}{HTML}{1A1A1A}
\definecolor{bordergray}{HTML}{E0E0E0}

\usepackage{fancyhdr}
\usepackage{titlesec}

\titleformat{\section}
  {\normalfont\large\bfseries\color{darkslate}}
  {\thesection}
  {0.8em}
  {}
\titlespacing*{\section}{0pt}{2ex plus 1ex minus .2ex}{1ex plus .2ex}

\titleformat{\subsection}
  {\normalfont\normalsize\bfseries\color{darkslate}}
  {\thesubsection}
  {0.6em}
  {}
\titlespacing*{\subsection}{0pt}{1.5ex plus 1ex minus .2ex}{0.6ex plus .2ex}

\renewenvironment{abstract}
 {
  \centerline{\large\bfseries ABSTRACT}\vspace{0.5em}
  \begin{quote}\small
 }
 {
  \end{quote}
 }

\usepackage[most]{tcolorbox}
\usepackage{booktabs}
\usepackage{colortbl}
\usepackage{tabularx}
\usepackage{caption}
\usepackage{amsmath,amssymb,amsfonts,mathtools}
\usepackage{hyperref}

\hypersetup{
  colorlinks=true,
  linkcolor=arxivblue,
  citecolor=arxivblue,
  urlcolor=arxivblue,
  pdftitle={$title$}
}

\DeclareCaptionFont{captionfont}{\small}
\DeclareCaptionFont{captiontitle}{\small\bfseries}
\providecommand{\tightlist}{%
  \setlength{\itemsep}{1pt}\setlength{\parskip}{0pt}}

\usepackage{longtable}
\usepackage{array}
\usepackage{calc}
\usepackage{graphicx}
\makeatletter
\def\maxwidth{\ifdim\Gin@nat@width>\linewidth\linewidth\else\Gin@nat@width\fi}
\def\maxheight{\ifdim\Gin@nat@height>\textheight\textheight\else\Gin@nat@height\fi}
\makeatother
\setkeys{Gin}{width=\maxwidth,height=\maxheight,keepaspectratio}

\renewcommand{\figurename}{Figure}
\renewcommand{\tablename}{Table}
\renewcommand{\contentsname}{Contents}

\PassOptionsToPackage{
  backend=biber,
  style=ieee,
  maxnames=6,
  minnames=1,
  doi=true,
  isbn=false,
  url=false
}{biblatex}

\color{darkslate}
\makeatletter
\@ifpackageloaded{tcolorbox}{}{\usepackage[skins,breakable]{tcolorbox}}
\@ifpackageloaded{fontawesome5}{}{\usepackage{fontawesome5}}
\definecolor{quarto-callout-color}{HTML}{909090}
\definecolor{quarto-callout-note-color}{HTML}{0758E5}
\definecolor{quarto-callout-important-color}{HTML}{CC1914}
\definecolor{quarto-callout-warning-color}{HTML}{EB9113}
\definecolor{quarto-callout-tip-color}{HTML}{00A047}
\definecolor{quarto-callout-caution-color}{HTML}{FC5300}
\definecolor{quarto-callout-color-frame}{HTML}{acacac}
\definecolor{quarto-callout-note-color-frame}{HTML}{4582ec}
\definecolor{quarto-callout-important-color-frame}{HTML}{d9534f}
\definecolor{quarto-callout-warning-color-frame}{HTML}{f0ad4e}
\definecolor{quarto-callout-tip-color-frame}{HTML}{02b875}
\definecolor{quarto-callout-caution-color-frame}{HTML}{fd7e14}
\makeatother
\makeatletter
\@ifpackageloaded{caption}{}{\usepackage{caption}}
\AtBeginDocument{%
\ifdefined\contentsname
  \renewcommand*\contentsname{Table of contents}
\else
  \newcommand\contentsname{Table of contents}
\fi
\ifdefined\listfigurename
  \renewcommand*\listfigurename{List of Figures}
\else
  \newcommand\listfigurename{List of Figures}
\fi
\ifdefined\listtablename
  \renewcommand*\listtablename{List of Tables}
\else
  \newcommand\listtablename{List of Tables}
\fi
\ifdefined\figurename
  \renewcommand*\figurename{Figure}
\else
  \newcommand\figurename{Figure}
\fi
\ifdefined\tablename
  \renewcommand*\tablename{Table}
\else
  \newcommand\tablename{Table}
\fi
}
\@ifpackageloaded{float}{}{\usepackage{float}}
\floatstyle{ruled}
\@ifundefined{c@chapter}{\newfloat{codelisting}{h}{lop}}{\newfloat{codelisting}{h}{lop}[chapter]}
\floatname{codelisting}{Listing}

\makeatother
\makeatletter
\@ifpackageloaded{caption}{}{\usepackage{caption}}
\@ifpackageloaded{subcaption}{}{\usepackage{subcaption}}
\makeatother
\usepackage{bookmark}
\IfFileExists{xurl.sty}{\usepackage{xurl}}{} 
\makeatletter
\@ifundefined{xmpquote}{\newcommand{\xmpquote}[1]{#1}}{}
\makeatother
\hypersetup{
  pdftitle={TAM-Chain: Multi-Scale Thyroid Cytology Classification via Absorbing Markov Chains and Shannon Entropy Uncertainty Quantification for False-Negative Suppression and Domain-Shift Adaptation},
  pdfauthor={Hai Pham Ngoc},
  pdfkeywords={\xmpquote{Absorbing Markov Chains, Shannon Entropy,
Thyroid Cytology, Optimal Stopping, Uncertainty Quantification, Domain
Shift Adaptation}},
  colorlinks=true,
  linkcolor={blue},
  filecolor={Maroon},
  citecolor={Blue},
  urlcolor={Blue},
  pdfcreator={LaTeX via pandoc}}

\title{TAM-Chain: Multi-Scale Thyroid Cytology Classification via
Absorbing Markov Chains and Shannon Entropy Uncertainty Quantification
for False-Negative Suppression and Domain-Shift Adaptation}
\author{Hai Pham Ngoc}
\date{2026-09-23}
\begin{document}
\thispagestyle{fancy}
\fancyhead[L]{\small\sffamily\color{gray} PREPRINT}
\fancyhead[R]{\small\sffamily\color{gray} 2026-09-23}

\vspace*{-0.5cm}

\begin{center}
  {\hrule height 1.5pt}
  \vspace{0.35cm}
  
  {\Large \bfseries \MakeUppercase{TAM-Chain: Multi-Scale Thyroid
Cytology Classification via Absorbing Markov Chains and Shannon Entropy
Uncertainty Quantification for False-Negative Suppression and
Domain-Shift Adaptation}} \\[0.2cm]
    
  \vspace{0.15cm}
  {\hrule height 1.5pt}
  
  \vspace{0.4cm}
  
  {\small \sffamily \bfseries \MakeUppercase{A Preprint}}
  
  \vspace{0.6cm}

      {\large \bfseries Hai Pham Ngoc} \\[0.15cm]
        {\small Faculty of Mathematics, Mechanics and Informatics, VNU
University of Science} \\[0.1cm]
        {\small \ttfamily harito.work@gmail.com} \\[0.2cm]
        {\small \color{gray} 2026-09-23}
  
\end{center}

\vspace{0.3cm}

\begin{abstract}
  \textbf{Background \& Problem:} Thyroid Fine-Needle Aspiration Biopsy
  (FNAB) cytology based on the Bethesda System plays a pivotal role in
  early thyroid cancer detection; however, deep learning approaches face
  substantial challenges regarding high false-negative rates and
  overconfidence under clinical domain shift.

  \textbf{Methods:} In this study, we propose \textbf{TAM-Chain}, a
  multi-scale (\(10\times, 20\times, 40\times\)) thyroid cytology
  classification framework leveraging \textbf{Absorbing Markov Chain}
  theory combined with Shannon Entropy-based Uncertainty Quantification.
  The framework dynamically models multi-magnification feature
  extraction as an absorbing stochastic process, enabling optimal
  stopping criteria and a human-in-the-loop referral mechanism to
  strictly suppress critical diagnostic errors.

  \textbf{Results:} Extensive evaluation on an internal test set
  (\(N=235\)) demonstrates a Macro F1 score of \textbf{\(0.9741\)} with
  an absolute False-Negative Rate (FNR) of \textbf{\(0.00\%\)}. On an
  independent external validation set (\(N=1015\)) presenting severe
  domain shift, TAM-Chain maintains superior stability and
  classification performance (\(\text{Macro F1} = 0.7026\)) by
  adaptively adjusting the expected stopping step and triggering
  specialist referrals, significantly outperforming single-magnification
  baselines.

  \textbf{Conclusion:} The TAM-Chain framework proves to be a highly
  effective, safe, and adaptable solution for digital pathology
  workflows, successfully harmonizing automated diagnostic efficiency
  with stringent biological safety.
\end{abstract}

\begin{center}
\end{center}
\vspace{0.5cm}


\pagestyle{fancy}
\pagenumbering{arabic}

\section{Introduction}\label{sec-introduction}

\subsection{Clinical Background and Challenges in Thyroid Cytology
Diagnosis}\label{sec-clinical-background}

Fine-Needle Aspiration Biopsy (FNAB) combined with The Bethesda System
for Reporting Thyroid Cytopathology (TBSRTC) serves as the gold standard
for clinical evaluation and management of thyroid nodules
(\citeproc{ref-pub.1194634019}{Negrelli et al. 2025};
\citeproc{ref-pub.1202783704}{Pham Ngoc et al. 2026}). In clinical
practice, three primary risk tiers dictate therapeutic strategies:

\begin{itemize}
\tightlist
\item
  \textbf{Bethesda II (B2 - Benign):} Indicates a benign condition
  manageable via routine clinical follow-up without surgical
  intervention.
\item
  \textbf{Bethesda V (B5 - Suspicious for Malignancy):} Warrants
  supplementary diagnostic testing or consideration of surgical
  resection.
\item
  \textbf{Bethesda VI (B6 - Malignant):} Mandates surgical intervention
  (\citeproc{ref-pub.1200302496}{Poyrazer and Erten 2026};
  \citeproc{ref-pub.1202783704}{Pham Ngoc et al. 2026}).
\end{itemize}

The paramount challenge for Medical Artificial Intelligence (Medical AI)
systems in this domain lies in controlling the \textbf{False Negative
Rate (FNR)} (\citeproc{ref-pub.1144320992}{Catak and Şahinbaş 2022}).
Misclassifying a patient with malignant cellular features (\(B5\) or
\(B6\)) as benign (\(B2\)) defers critical surgical intervention,
incurring catastrophic consequences for patient health and clinical
outcomes (\citeproc{ref-pub.1144320992}{Catak and Şahinbaş 2022};
\citeproc{ref-pub.1194634019}{Negrelli et al. 2025}). Consequently,
ensuring clinical safety with \(\text{FNR} \approx 0\%\) via trustworthy
oversight mechanisms represents an indispensable prerequisite for
deploying automated diagnostic models into clinical workflows
(\citeproc{ref-pub.1189327682}{Rashidisabet et al. 2025};
\citeproc{ref-pub.1202318764}{Patel et al. 2026}).

\begin{tcolorbox}[enhanced jigsaw, arc=.35mm, bottomrule=.15mm, bottomtitle=1mm, breakable, colback=white, colbacktitle=quarto-callout-warning-color!10!white, colframe=quarto-callout-warning-color-frame, coltitle=black, left=2mm, leftrule=.75mm, opacityback=0, opacitybacktitle=0.6, rightrule=.15mm, title=\textcolor{quarto-callout-warning-color}{\faExclamationTriangle}\hspace{0.5em}{Stringent Clinical Safety Requirements}, titlerule=0mm, toprule=.15mm, toptitle=1mm]

In thyroid cancer screening, diagnostic misclassifications from
\(B5/B6 \to B2\) (False Negatives) are strictly unacceptable within
clinical protocols. AI models must not only achieve high overall
accuracy but also possess reliable referral/deferral mechanisms to
delegate high-risk or ambiguous cases to expert pathologists.

\end{tcolorbox}

\subsection{Limitations of Single-Scale Deep Learning Models and Domain
Shift}\label{sec-limitations}

Although Deep Learning has driven landmark advancements in medical image
analysis, deploying conventional machine learning models for multi-scale
cytological diagnosis remains constrained by two major technical
bottlenecks:

\begin{itemize}
\tightlist
\item
  \textbf{Multi-Scale Computational and Contextual Trade-Off:} Analyzing
  Whole Slide Images (WSIs) requires examination across multiple
  magnifications (\citeproc{ref-pub.1202783704}{Pham Ngoc et al. 2026}).
  Inspection at low magnification (\(10\times\)) provides broad spatial
  architectural context but risks overlooking subtle nuclear cytological
  details; conversely, exhaustively processing images at high
  magnification (\(40\times\)) incurs substantial computational overhead
  and memory burdens, while elevating the risk of cellular context loss
  (\citeproc{ref-pub.1193389571}{Rui et al. 2026};
  \citeproc{ref-pub.1202783704}{Pham Ngoc et al. 2026}).
\item
  \textbf{Model Overconfidence Under Domain Shift:} When deployed on
  external validation datasets characterized by distributional
  shifts---stemming from inter-site variations in staining techniques,
  scanner models, or slide preparation protocols---deep neural networks
  frequently exhibit performance degradation and issue egregious
  diagnostic errors with undesirably high confidence (Overconfident
  Errors) (\citeproc{ref-pub.1164408629}{Araújo et al. 2023};
  \citeproc{ref-pub.1189327682}{Rashidisabet et al. 2025};
  \citeproc{ref-pub.1195220495}{Hossain et al. 2026}).
\end{itemize}

\subsection{Proposed Framework: TAM-Chain}\label{sec-proposed-framework}

To rigorously address these challenges, this study introduces
\textbf{TAM-Chain} (\textbf{T}hyroid \textbf{A}bsorbing \textbf{M}arkov
\textbf{Chain})---a multi-scale, dynamic stochastic stopping framework
grounded in \textbf{Absorbing Markov Chain} theory
(\citeproc{ref-pub.1033238727}{Saul I. Gass 2013}) coupled with
\textbf{Shannon Entropy-based Uncertainty Quantification (UQ)}
(\citeproc{ref-pub.1141515606}{Kang et al. 2021};
\citeproc{ref-pub.1172313676}{Huang et al. 2024}).

\begin{figure}

\centering{

\includegraphics[width=8in,height=8.5in]{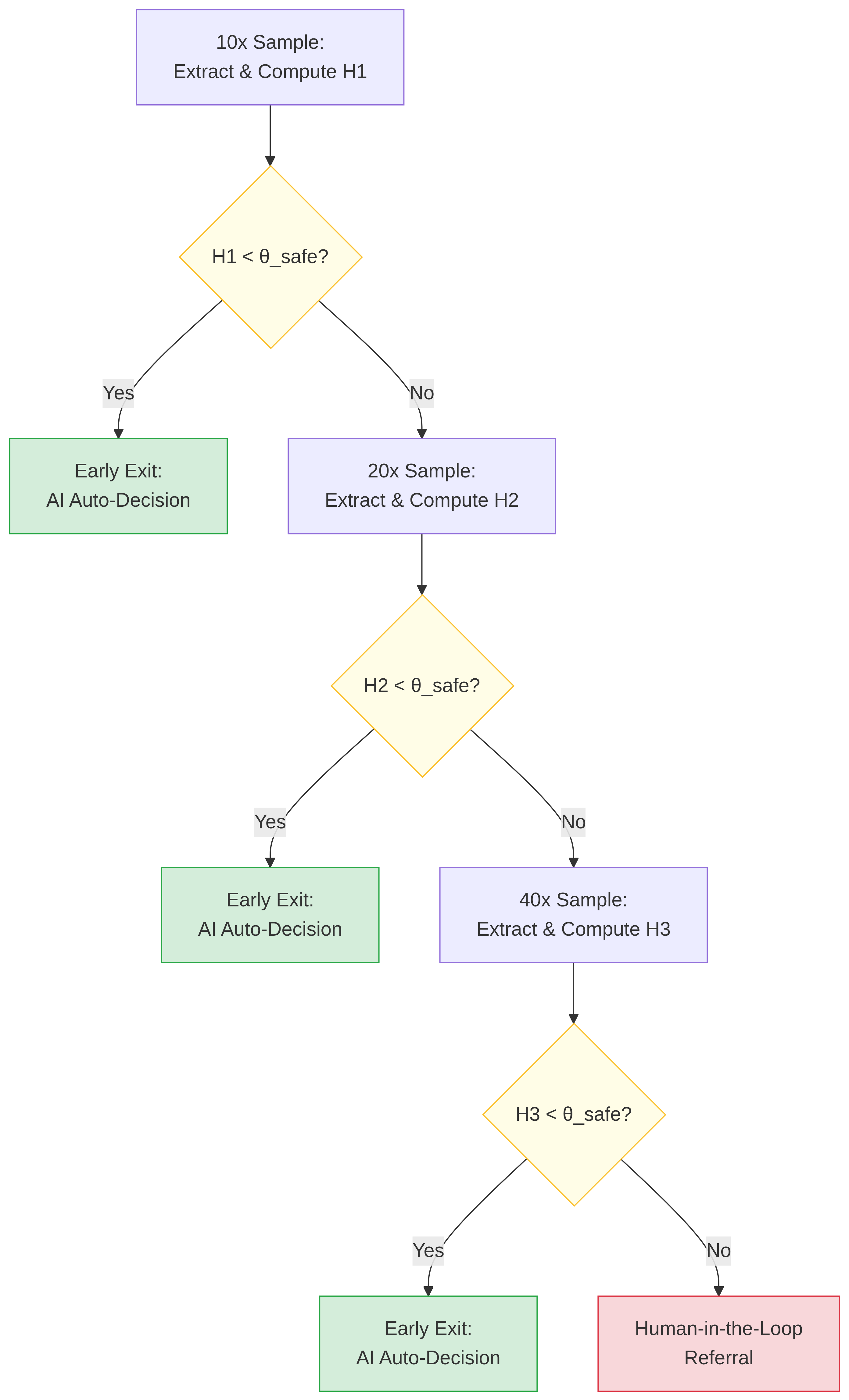}

}

\caption{\label{fig-tam-chain-flow}Stochastic state transition workflow
and dynamic early stopping in the TAM-Chain framework}

\end{figure}

The decision flow of the proposed TAM-Chain framework is illustrated in
Figure~\ref{fig-tam-chain-flow}.

The core operational mechanisms of TAM-Chain comprise:

\begin{itemize}
\tightlist
\item
  \textbf{Dynamic Early Stopping Mechanism:} Optimizes the stopping step
  \(\tau\) by continuously monitoring uncertainty \(H(p_t)\) across zoom
  hierarchies (\(10\times \to 20\times \to 40\times\)). When uncertainty
  falls below the safety threshold \(\theta_{\text{safe}}\), the Markov
  chain gets absorbed into state \(A_1\), where the AI finalized the
  diagnosis automatically without incurring unnecessary
  high-magnification computational costs
  (\citeproc{ref-pub.1145665340}{Zhang et al. 2022};
  \citeproc{ref-pub.1193389571}{Rui et al. 2026}).
\item
  \textbf{Human-in-the-Loop Referral Mechanism:} For ambiguous, noisy,
  or domain-shifted specimens where uncertainty remains elevated even
  after observing all three magnification levels, the chain
  automatically transitions into absorbing state \(A_2\). This state
  functions as a safety trigger, deferring the case to expert
  pathologists for manual evaluation (Human-in-the-Loop)
  (\citeproc{ref-pub.1141515606}{Kang et al. 2021};
  \citeproc{ref-pub.1144320992}{Catak and Şahinbaş 2022};
  \citeproc{ref-pub.1199646874}{Aktas et al. 2026}), thereby mitigating
  hazardous overconfident errors
  (\citeproc{ref-pub.1189327682}{Rashidisabet et al. 2025};
  \citeproc{ref-pub.1195220495}{Hossain et al. 2026}).
\item
  \textbf{Multi-Scale Ensemble Strategy (\texttt{AvgProbs}):} For
  challenging cases routed to \(A_2\), the framework aggregates
  multi-scale diagnostic information by computing the average
  probability matrix \texttt{AvgProbs}, ensuring maximum stability and
  diagnostic robustness prior to specialist referral
  (\citeproc{ref-pub.1196199978}{Rossi et al. 2025};
  \citeproc{ref-pub.1202783704}{Pham Ngoc et al. 2026}).
\end{itemize}

\subsection{Core Contributions}\label{sec-contributions}

The primary contributions of this study are summarized as follows:

\begin{itemize}
\tightlist
\item
  \textbf{Formulation of a Multi-Scale Stochastic Model via Absorbing
  Markov Chains:} We construct a rigorous mathematical framework
  integrating Absorbing Markov Chains
  (\citeproc{ref-pub.1033238727}{Saul I. Gass 2013}) and Shannon Entropy
  (\citeproc{ref-pub.1141515606}{Kang et al. 2021}), facilitating
  dynamic early stopping at appropriate magnification tiers. This
  significantly reduces computational costs
  (\(\mathbb{E}[\tau] \approx 2.0181\)) while preserving top-tier
  diagnostic fidelity.
\item
  \textbf{Absolute Suppression of False Negatives on Internal
  Validation:} Achieves a Macro F1 score of \(0.9741\) on the internal
  test set (\(N=235\)) with an absolute False Negative Rate
  (\(\text{FNR} = 0.00\%\)) within the automated decision subset
  (\(A_1\)), effectively eliminating the risk of misdiagnosing malignant
  \(B5/B6\) tumors.
\item
  \textbf{Adaptive Safety and Resilience Under Domain Shift (External
  Validation):} Independent evaluation on an external validation cohort
  (\(N=1015\)) demonstrates that the Markov chain adaptively responds to
  out-of-domain data by extending average observation steps
  (\(\mathbb{E}[\tau]: 2.0181 \to 2.4944\)) and expanding human referral
  rates (\(P(A_2): 44.65\% \to 66.84\%\)), serving as a robust safeguard
  for clinical deployment (\citeproc{ref-pub.1189327682}{Rashidisabet et
  al. 2025}; \citeproc{ref-pub.1199646874}{Aktas et al. 2026}).
\item
  \textbf{Rigorous Variance Analysis and Statistical Testing:} Through
  1,000 Bootstrap resampling iterations combined with non-parametric
  Friedman tests (\(p < 0.001\)), we demonstrate statistically
  significant superiority and minimal variance (\(\sigma^2 = 0.000110\))
  for the \texttt{AvgProbs} ensemble strategy compared to single-scale
  baselines.
\end{itemize}

\begin{tcolorbox}[enhanced jigsaw, arc=.35mm, bottomrule=.15mm, bottomtitle=1mm, breakable, colback=white, colbacktitle=quarto-callout-note-color!10!white, colframe=quarto-callout-note-color-frame, coltitle=black, left=2mm, leftrule=.75mm, opacityback=0, opacitybacktitle=0.6, rightrule=.15mm, title=\textcolor{quarto-callout-note-color}{\faInfo}\hspace{0.5em}{Paper Organization}, titlerule=0mm, toprule=.15mm, toptitle=1mm]

The remainder of this manuscript is organized as follows:
Section~\ref{sec-introduction} establishes the clinical background and
study motivations; Section~\ref{sec-related-works} reviews related
literature on uncertainty quantification and Markovian models in digital
pathology; Section~\ref{sec-methodology} details the mathematical
formulation of the TAM-Chain framework; Section~\ref{sec-exp-setup}
outlines the experimental setup and dataset composition;
Section~\ref{sec-results} presents quantitative findings and statistical
analyses; Section~\ref{sec-discussion} provides an in-depth clinical
discussion, adaptive mechanisms, and study limitations; and finally,
Section~\ref{sec-conclusion} concludes the paper and discusses future
research directions.

\end{tcolorbox}

\section{Related Work}\label{sec-related-works}

\subsection{Thyroid Cytology Diagnosis and Limitations of Single-Scale
Deep Learning}\label{sec-related-thyroid}

Fine-Needle Aspiration Biopsy (FNAB) plays a pivotal role in thyroid
cancer screening protocols governed by The Bethesda System for Reporting
Thyroid Cytopathology (\citeproc{ref-pub.1194634019}{Negrelli et al.
2025}; \citeproc{ref-pub.1202783704}{Pham Ngoc et al. 2026}). The
application of Deep Learning (DL) models to automated thyroid cytology
classification has achieved substantial progress in recent years
(\citeproc{ref-pub.1200302496}{Poyrazer and Erten 2026}). Early studies
primarily focused on deploying Convolutional Neural Networks (CNNs),
such as ResNet or EfficientNet, on image patches extracted at a fixed
single magnification level (\citeproc{ref-pub.1200302496}{Poyrazer and
Erten 2026}; \citeproc{ref-pub.1202783704}{Pham Ngoc et al. 2026}).

However, single-scale approaches exhibit severe constraints in clinical
practice:

\begin{itemize}
\tightlist
\item
  \textbf{Low-Magnification Inspection (\(10\times\)):} Models preserve
  a global architectural overview of cell cluster density and spatial
  distribution but lack critical nuclear micro-features (e.g., nuclear
  grooves, intranuclear inclusions, or psammoma bodies).
\item
  \textbf{High-Magnification Inspection (\(40\times\)):} Models capture
  detailed nuclear morphological characteristics but suffer from
  cellular context loss, while concurrently escalating computational
  memory overhead when processing Whole Slide Images (WSIs)
  (\citeproc{ref-pub.1193389571}{Rui et al. 2026};
  \citeproc{ref-pub.1202783704}{Pham Ngoc et al. 2026}).
\end{itemize}

\subsection{Uncertainty Quantification and Human-in-the-Loop
Safeguards}\label{sec-related-uq}

In safety-critical Medical AI applications, issuing accurate predictions
alone is insufficient; models must possess ``awareness of their own
ignorance'' (Uncertainty Awareness) (\citeproc{ref-pub.1141515606}{Kang
et al. 2021}; \citeproc{ref-pub.1172313676}{Huang et al. 2024}).
Diagnostic uncertainty in medical imaging is broadly categorized into
two forms:

\begin{itemize}
\tightlist
\item
  \textbf{Aleatoric Uncertainty:} Arises from intrinsic data noise,
  non-uniform staining, or out-of-focus and overlapping cellular
  artifacts.
\item
  \textbf{Epistemic Uncertainty:} Stems from model ignorance when
  encountering out-of-distribution (OOD) data absent from the training
  set.
\end{itemize}

To quantify uncertainty, Shannon Entropy \(H(p)\) computed over the
output probability distribution is widely adopted due to its
mathematical rigor and minimal computational overhead
(\citeproc{ref-pub.1141515606}{Kang et al. 2021};
\citeproc{ref-pub.1193389571}{Rui et al. 2026}). When integrated into a
Human-in-the-Loop Referral mechanism, predictions exhibiting uncertainty
above a predefined safety threshold are flagged and deferred to expert
pathologists for manual review (\citeproc{ref-pub.1144320992}{Catak and
Şahinbaş 2022}; \citeproc{ref-pub.1199646874}{Aktas et al. 2026}). This
strategy effectively suppresses the False Negative Rate (FNR),
safeguarding patients against misdiagnosed malignancies
(\citeproc{ref-pub.1144320992}{Catak and Şahinbaş 2022};
\citeproc{ref-pub.1189327682}{Rashidisabet et al. 2025}).

\subsection{Domain Shift and the Challenge of Model
Overconfidence}\label{sec-related-domain-shift}

A major bottleneck hindering the clinical deployment of Medical AI is
performance degradation on external validation cohorts---a phenomenon
known as \textbf{Domain Shift} (\citeproc{ref-pub.1164408629}{Araújo et
al. 2023}; \citeproc{ref-pub.1189327682}{Rashidisabet et al. 2025}).
Cross-institutional variations in WSI scanner hardware, slide
preparation protocols, section thickness, and Giemsa/Papanicolaou
staining techniques induce significant feature distribution shifts
(\citeproc{ref-pub.1194634019}{Negrelli et al. 2025};
\citeproc{ref-pub.1195220495}{Hossain et al. 2026}).

Under domain shift, conventional deep neural networks frequently suffer
from \textbf{Model Overconfidence}: outputting erroneous diagnostic
predictions with undesirably high confidence scores (\(> 95\%\))
(\citeproc{ref-pub.1189327682}{Rashidisabet et al. 2025};
\citeproc{ref-pub.1195220495}{Hossain et al. 2026}). Consequently, naive
confidence-gating mechanisms based on maximum softmax probability fail
to detect hazardous diagnostic errors.

\subsection{Absorbing Markov Chains in Multi-Tiered Decision
Optimization}\label{sec-related-markov}

The theory of \textbf{Absorbing Markov Chains (AMCs)} provides a
mathematically rigorous formulation for modeling stochastic processes
featuring terminal absorbing states (\citeproc{ref-pub.1033238727}{Saul
I. Gass 2013}). In operations research and decision optimization, AMCs
facilitate the exact analytical computation of expected observation
steps \(\mathbb{E}[\tau]\) and absorption probabilities via the
Fundamental Matrix \(N = (I - Q)^{-1}\)
(\citeproc{ref-pub.1033238727}{Saul I. Gass 2013}).

Although Markovian models have been applied to treatment planning and
disease progression forecasting (\citeproc{ref-pub.1033238727}{Saul I.
Gass 2013}), coupling Absorbing Markov Chains with \textbf{Shannon
Entropy} as dynamic state-transition triggers across microscopic optical
magnifications (\(10\times \to 20\times \to 40\times\)) represents a
novel research paradigm.

\subsection{Comparative Synthesis and Research
Gap}\label{sec-related-gap}

Table~\ref{tbl-related-works} presents a systematic comparison between
existing methodologies and the proposed \textbf{TAM-Chain} framework.

\begin{longtable}[]{@{}
  >{\raggedright\arraybackslash}p{(\linewidth - 10\tabcolsep) * \real{0.1379}}
  >{\centering\arraybackslash}p{(\linewidth - 10\tabcolsep) * \real{0.1724}}
  >{\centering\arraybackslash}p{(\linewidth - 10\tabcolsep) * \real{0.1724}}
  >{\centering\arraybackslash}p{(\linewidth - 10\tabcolsep) * \real{0.1724}}
  >{\centering\arraybackslash}p{(\linewidth - 10\tabcolsep) * \real{0.1724}}
  >{\centering\arraybackslash}p{(\linewidth - 10\tabcolsep) * \real{0.1724}}@{}}
\caption{Comparative analysis between related works and the proposed
TAM-Chain framework}\label{tbl-related-works}\tabularnewline
\toprule\noalign{}
\begin{minipage}[b]{\linewidth}\raggedright
Methodology
\end{minipage} & \begin{minipage}[b]{\linewidth}\centering
Multi-Scale Handling
\end{minipage} & \begin{minipage}[b]{\linewidth}\centering
Uncertainty Quantification
\end{minipage} & \begin{minipage}[b]{\linewidth}\centering
FNR Control
\end{minipage} & \begin{minipage}[b]{\linewidth}\centering
Domain Shift Adaptability
\end{minipage} & \begin{minipage}[b]{\linewidth}\centering
Computational Efficiency
\end{minipage} \\
\midrule\noalign{}
\endfirsthead
\toprule\noalign{}
\begin{minipage}[b]{\linewidth}\raggedright
Methodology
\end{minipage} & \begin{minipage}[b]{\linewidth}\centering
Multi-Scale Handling
\end{minipage} & \begin{minipage}[b]{\linewidth}\centering
Uncertainty Quantification
\end{minipage} & \begin{minipage}[b]{\linewidth}\centering
FNR Control
\end{minipage} & \begin{minipage}[b]{\linewidth}\centering
Domain Shift Adaptability
\end{minipage} & \begin{minipage}[b]{\linewidth}\centering
Computational Efficiency
\end{minipage} \\
\midrule\noalign{}
\endhead
\bottomrule\noalign{}
\endlastfoot
Single-Scale CNN (\citeproc{ref-pub.1200302496}{Poyrazer and Erten
2026}) & None & None & None & None & Low \\
Static Multi-Scale Fusion (\citeproc{ref-pub.1202783704}{Pham Ngoc et
al. 2026}) & Static & None & None & Poor & Very Low \\
Softmax Thresholding (\citeproc{ref-pub.1141515606}{Kang et al. 2021}) &
None & Superficial & Partial & Prone to Overconfidence & Moderate \\
Uncertainty Triage (\citeproc{ref-pub.1199646874}{Aktas et al. 2026}) &
None & Applied & Strong & Moderate & Moderate \\
\textbf{TAM-Chain (Proposed)} & \textbf{Dynamic
(\(10\times \to 40\times\))} & \textbf{Shannon Entropy} &
\textbf{\(\text{FNR} = 0\%\)} & \textbf{Self-adapting
\(\mathbb{E}[\tau]\) \& \(P(A_2)\)} & \textbf{Optimal
(\(\mathbb{E}[\tau] \approx 2.01\))} \\
\end{longtable}

\begin{tcolorbox}[enhanced jigsaw, arc=.35mm, bottomrule=.15mm, bottomtitle=1mm, breakable, colback=white, colbacktitle=quarto-callout-note-color!10!white, colframe=quarto-callout-note-color-frame, coltitle=black, left=2mm, leftrule=.75mm, opacityback=0, opacitybacktitle=0.6, rightrule=.15mm, title=\textcolor{quarto-callout-note-color}{\faInfo}\hspace{0.5em}{Research Gap Addressed}, titlerule=0mm, toprule=.15mm, toptitle=1mm]

TAM-Chain bridges key literature gaps by unifying: (1) Computational
efficiency via dynamic stochastic early stopping, (2) Clinical safety
guarantees through automated human expert referrals (\(A_2\)), and (3)
Inherent adaptability under domain shift without requiring model
retraining.

\end{tcolorbox}

\section{Methodology}\label{sec-methodology}

\subsection{Architectural Overview of the TAM-Chain
Framework}\label{sec-methodology-overview}

The \textbf{TAM-Chain} (\textbf{T}hyroid \textbf{A}bsorbing
\textbf{M}arkov \textbf{Chain}) framework is designed as a multi-tier
stochastic diagnostic pipeline that integrates Deep Learning for
cellular morphological feature extraction with Absorbing Markov Chains
for dynamic early stopping and safe referral.

Given a Fine-Needle Aspiration Biopsy (FNAB) cytological image sample,
the system processes data sequentially across ascending magnification
levels: \(10\times \to 20\times \to 40\times\). At each magnification
tier \(t \in \{1, 2, 3\}\), a deep neural network extracts feature
vectors and outputs a predicted probability distribution
\(p_t = [p_t^{(B2)}, p_t^{(B5)}, p_t^{(B6)}]^T\) corresponding to the
three representative Bethesda categories.

\subsection{\texorpdfstring{Backbone Architecture and Model Deployment
(\texttt{ThyroidEffi})}{Backbone Architecture and Model Deployment (ThyroidEffi)}}\label{sec-backbone-implementation}

To ensure precise cellular morphological feature extraction across
varied magnification levels, TAM-Chain utilizes a specialized backbone
model designated as \textbf{\texttt{ThyroidEffi\ 1.0\ Base}}.

\subsubsection{Backbone Architecture and
Training}\label{backbone-architecture-and-training}

\begin{itemize}
\tightlist
\item
  \textbf{Architectural Foundation:} Built upon optimizations of the
  high-performance EfficientNet family, integrating
  Squeeze-and-Excitation (SE) blocks to focus on critical
  micro-architectural regions of thyroid cell nuclei.
\item
  \textbf{Loss Function and Optimization:} The network is trained on the
  training set using Weighted Cross-Entropy Loss to address class
  imbalance across Bethesda groups (\(B2, B5, B6\)). Optimization is
  conducted via AdamW with an initial learning rate of \(10^{-4}\) and a
  Cosine Annealing decay schedule.
\item
  \textbf{Data Augmentation:} Detailed data augmentation protocols
  follow the methodologies established in
  (\citeproc{ref-pub.1202783704}{Pham Ngoc et al. 2026}).
\end{itemize}

\subsubsection{Real-Time Deployment via
TorchScript}\label{real-time-deployment-via-torchscript}

\begin{itemize}
\tightlist
\item
  To seamlessly integrate with the Absorbing Markov Chain framework and
  optimize real-time inference speed, the trained backbone weights are
  compiled and exported using \textbf{TorchScript}.
\item
  Executing inference via TorchScript eliminates Python interpreter
  overhead, enabling rapid extraction of probability distributions
  \(p_t\) at each transition step \(10\times \to 20\times \to 40\times\)
  prior to evaluation by the Shannon Entropy function \(H(p_t)\).
\end{itemize}

\subsection{Multi-Scale Patch Extraction and
Preprocessing}\label{sec-patch-extraction}

To provide standardized inputs for the TAM-Chain framework across the
three magnification tiers (\(10\times, 20\times, 40\times\)), Whole
Slide Image (WSI) preprocessing is executed according to the following
protocol:

\subsubsection{Specimen Normalization and Region of Interest (ROI)
Segmentation}\label{specimen-normalization-and-region-of-interest-roi-segmentation}

\begin{itemize}
\tightlist
\item
  FNAB slides acquired from digital scanners are partitioned into
  non-overlapping or controlled-overlapping patch grids at base
  resolution.
\item
  To mitigate contextual information loss during localized patch
  cropping and prevent nuclear detail degradation caused by resizing,
  the system constructs multi-scale image pyramids corresponding to
  three objective lens magnifications:

  \begin{itemize}
  \tightlist
  \item
    \textbf{\(10\times\) Tier (\(S_1\)):} Provides a wide field-of-view
    (FOV), preserving cellular cluster density, spatial arrangement, and
    overall thyroid papillary architecture.
  \item
    \textbf{\(20\times\) Tier (\(S_2\)):} Balances local context with
    intermediate resolution, facilitating clearer identification of
    epithelial cell clusters.
  \item
    \textbf{\(40\times\) Tier (\(S_3\)):} Zoom-in to microscopic
    resolution, extracting fine nuclear details (such as psammoma
    bodies, nuclear grooves, nuclear membrane irregularities, and pale
    chromatin)---crucial determinants for Bethesda \(B5/B6\)
    classification.
  \end{itemize}
\end{itemize}

\subsubsection{Multi-Scale Feature
Alignment}\label{multi-scale-feature-alignment}

\begin{itemize}
\tightlist
\item
  Corresponding patch regions sharing identical spatial center
  coordinates on the WSI are extracted simultaneously across all three
  magnification tiers.
\item
  Each patch at every level is passed through the feature extraction
  backbone (\texttt{ThyroidEffi\ 1.0\ Base}) to generate predicted
  probability vectors \(p_{10\times}, p_{20\times}, p_{40\times}\),
  serving as inputs for Shannon Entropy \(H(p_t)\) computation and state
  control within the Absorbing Markov Chain.
\end{itemize}

\subsection{Uncertainty Quantification via Shannon
Entropy}\label{sec-methodology-entropy}

To quantify prediction uncertainty (Uncertainty Quantification - UQ) at
magnification tier \(t\), we employ \textbf{Shannon Entropy} \(H(p_t)\)
(\citeproc{ref-pub.1141515606}{Kang et al. 2021};
\citeproc{ref-pub.1172313676}{Huang et al. 2024}):

\[H(p_t) = -\sum_{c \in \{B2, B5, B6\}} p_t^{(c)} \log_2 \left( p_t^{(c)} \right)\]
\{\#eq-shannon-entropy\}

where \(p_t^{(c)}\) represents the predicted probability for class \(c\)
at tier \(t\).

\begin{itemize}
\tightlist
\item
  When the model is highly confident in a single class prediction,
  \(H(p_t) \to 0\).
\item
  When the model exhibits high ambiguity among classes (particularly
  under noise or domain shift), \(H(p_t)\) approaches its theoretical
  maximum of \(\log_2(3) \approx 1.585\) bits.
\end{itemize}

The value \(H(p_t)\) is evaluated directly against a predefined safety
threshold \(\theta_{\text{safe}}\), optimized on the validation set, to
govern state transitions within the Markov chain.

\subsection{Absorbing Markov Chain
Formulation}\label{sec-methodology-markov-chain}

\subsubsection{State Space Definition}\label{sec-methodology-states}

We formulate the multi-magnification diagnostic process as a
discrete-time Markov chain \(\{X_t\}_{t=1}^3\) defined over the state
space \(S = T \cup A\):

\begin{itemize}
\tightlist
\item
  \textbf{Transient State Set (\(T\)):}

  \begin{itemize}
  \tightlist
  \item
    \(S_1\): Inspection at \(10\times\) magnification.
  \item
    \(S_2\): Inspection at \(20\times\) magnification.
  \item
    \(S_3\): Inspection at \(40\times\) magnification.
  \end{itemize}
\item
  \textbf{Absorbing State Set (\(A\)):}

  \begin{itemize}
  \tightlist
  \item
    \(A_1\) (\textbf{AI Self-Diagnosis}): Early exit where the AI
    automatically finalizes the diagnosis when uncertainty
    \(H(p_t) < \theta_{\text{safe}}\).
  \item
    \(A_2\) (\textbf{Human-in-the-Loop Referral}): Case deferral to
    expert pathologists when uncertainty remains elevated after three
    inspection tiers (\(H(p_3) \ge \theta_{\text{safe}}\)).
  \end{itemize}
\end{itemize}

\subsubsection{Transition Probability
Matrix}\label{sec-methodology-transition-matrix}

The transition probability matrix \(P\) of the Markov chain is expressed
in canonical form (\citeproc{ref-pub.1033238727}{Saul I. Gass 2013}):

\[P = \begin{pmatrix} Q & R \\ \mathbf{0} & I \end{pmatrix}\]
\{\#eq-canonical-p\}

where \(Q \in \mathbb{R}^{3 \times 3}\) models transition probabilities
among transient states, \(R \in \mathbb{R}^{3 \times 2}\) models
transition probabilities from transient states to absorbing states,
\(\mathbf{0}\) is a zero matrix, and \(I \in \mathbb{R}^{2 \times 2}\)
is the identity matrix.

The explicit components of matrix \(P\) are governed by the stopping
rules:

\[P = \begin{pmatrix} 0 & q_1 & 0 & r_{1,A1} & 0 \\ 0 & 0 & q_2 & r_{2,A1} & 0 \\ 0 & 0 & 0 & r_{3,A1} & r_{3,A2} \\ 0 & 0 & 0 & 1 & 0 \\ 0 & 0 & 0 & 0 & 1 \end{pmatrix}\]
\{\#eq-full-transition-matrix\}

with transition probabilities empirically derived from validation data:

\begin{itemize}
\tightlist
\item
  \(r_{t,A1} = P(H(p_t) < \theta_{\text{safe}} \mid S_t)\) represents
  the proportion of cases reaching the safety criterion at tier \(t\).
\item
  \(q_t = 1 - r_{t,A1}\) represents the proportion of cases requiring
  escalation to magnification tier \(t+1\).
\item
  At state \(S_3\), if \(H(p_3) \ge \theta_{\text{safe}}\), the case is
  obligatorily routed to absorbing state \(A_2\) with probability
  \(r_{3,A2} = 1 - r_{3,A1}\).
\end{itemize}

\subsubsection{\texorpdfstring{Fundamental Matrix Analysis and Expected
Stopping Step
\(\mathbb{E}[\tau]\)}{Fundamental Matrix Analysis and Expected Stopping Step \textbackslash mathbb\{E\}{[}\textbackslash tau{]}}}\label{sec-methodology-fundamental-matrix}

According to Absorbing Markov Chain theory
(\citeproc{ref-pub.1033238727}{Saul I. Gass 2013}), the
\textbf{Fundamental Matrix} \(N\) denotes the expected number of steps
the chain spends in transient state \(S_j\) prior to absorption, given
initial state \(S_i\):

\[N = (I - Q)^{-1} = \sum_{k=0}^{\infty} Q^k\]
\{\#eq-fundamental-matrix\}

Since \(Q\) is strictly upper triangular, \(N\) can be analytically
computed in closed form:

\[N = \begin{pmatrix} 1 & q_1 & q_1 q_2 \\ 0 & 1 & q_2 \\ 0 & 0 & 1 \end{pmatrix}\]
\{\#eq-explicit-n\}

The \textbf{expected observation steps \(\mathbb{E}[\tau]\)} (reflecting
computational cost and the average magnification depth per case)
starting from \(S_1\) is obtained by summing the first row of \(N\):

\begin{equation}\protect\phantomsection\label{eq-expected-time}{
\mathbb{E}[\tau] = \sum_{j=1}^3 N_{1,j} = 1 + q_1 + q_1 q_2
}\end{equation}

The \textbf{absorption probability matrix
\(B \in \mathbb{R}^{3 \times 2}\)} indicates the probability that a
chain starting in \(S_i\) eventually terminates in absorbing state
\(A_j\):

\begin{equation}\protect\phantomsection\label{eq-absorption-probs}{
B = N \cdot R
}\end{equation}

The probability of early automated AI decision \(P(A_1)\) and the
probability of expert pathologist referral \(P(A_2)\) starting from
\(S_1\) are derived as:

\begin{equation}\protect\phantomsection\label{eq-prob-a1}{
P(A_1) = B_{1,1} = r_{1,A1} + q_1 r_{2,A1} + q_1 q_2 r_{3,A1}
}\end{equation}

\begin{equation}\protect\phantomsection\label{eq-prob-a2}{
P(A_2) = B_{1,2} = q_1 q_2 r_{3,A2} = 1 - P(A_1)
}\end{equation}

\begin{tcolorbox}[enhanced jigsaw, arc=.35mm, bottomrule=.15mm, bottomtitle=1mm, breakable, colback=white, colbacktitle=quarto-callout-note-color!10!white, colframe=quarto-callout-note-color-frame, coltitle=black, left=2mm, leftrule=.75mm, opacityback=0, opacitybacktitle=0.6, rightrule=.15mm, title=\textcolor{quarto-callout-note-color}{\faInfo}\hspace{0.5em}{Clinical and Computational Significance of Markov Metrics:}, titlerule=0mm, toprule=.15mm, toptitle=1mm]

\begin{itemize}
\tightlist
\item
  \textbf{Matrix \(N\) and \(\mathbb{E}[\tau]\)} measure
  \emph{computational efficiency}: Lower \(\mathbb{E}[\tau]\) values
  indicate greater computational savings by terminating prematurely at
  lower optical magnifications.
\item
  \textbf{Matrix \(B\) and probabilities \(\{P(A_1), P(A_2)\}\)} measure
  \emph{safety and automation trade-offs}: \(P(A_1)\) reflects the
  degree of AI autonomy, while \(P(A_2)\) acts as a safety mechanism,
  routing high-uncertainty cases to expert pathologists to maximize
  clinical safety.
\end{itemize}

\end{tcolorbox}

\subsection{\texorpdfstring{Multi-Scale Probability Aggregation Strategy
(\texttt{AvgProbs})}{Multi-Scale Probability Aggregation Strategy (AvgProbs)}}\label{sec-methodology-avgprobs}

For challenging cases absorbed into \(A_2\) (requiring supplementary
expert pathologist evaluation), the framework provides an optimized
probability recommendation to minimize context loss. The multi-scale
probability aggregation strategy (\texttt{AvgProbs}) computes the
unweighted arithmetic mean of the predicted probability matrices across
all three magnification tiers (\citeproc{ref-pub.1196199978}{Rossi et
al. 2025}; \citeproc{ref-pub.1202783704}{Pham Ngoc et al. 2026}):

\begin{equation}\protect\phantomsection\label{eq-avgprobs}{
\bar{p}_{A2} = \frac{1}{3} \sum_{t=1}^3 p_t = \frac{p_{10\times} + p_{20\times} + p_{40\times}}{3}
}\end{equation}

The final decision recommendation for set \(A_2\) (when AI guidance is
referenced) corresponds to the class maximizing the averaged
probability:

\begin{equation}\protect\phantomsection\label{eq-arg-max}{
\hat{y}_{A2} = \arg\max_{c \in \{B2, B5, B6\}} \bar{p}_{A2}^{(c)}
}\end{equation}

The \texttt{AvgProbs} strategy effectively smooths localized
high-magnification artifacts at \(40\times\) while incorporating spatial
context from \(10\times\), achieving minimal prediction variance and
enhanced stability under domain shift
(\citeproc{ref-pub.1189327682}{Rashidisabet et al. 2025};
\citeproc{ref-pub.1202783704}{Pham Ngoc et al. 2026}). In addition, four
baseline aggregation methods were implemented for comparative
evaluation, as detailed in Section~\ref{sec-baselines}.

\section{Experimental Setup}\label{sec-exp-setup}

\subsection{Datasets and Cohort Partitioning}\label{sec-datasets}

This study utilizes two independent Fine-Needle Aspiration Biopsy (FNAB)
thyroid cytology datasets collected from our previous work
(\citeproc{ref-pub.1202783704}{Pham Ngoc et al. 2026}):

\begin{itemize}
\tightlist
\item
  \textbf{Internal Dataset (108 Military Central Hospital):} Comprises
  \(1,804\) patients, partitioned at the patient level into three
  subsets:

  \begin{itemize}
  \tightlist
  \item
    \emph{Training Set:} Used to train the
    \texttt{ThyroidEffi\ 1.0\ Base} feature extraction backbone.
  \item
    \emph{Validation Set (\(N = 236\)):} Dedicated to calibrating the
    entropy safety threshold \(\theta_{\text{safe}}\).
  \item
    \emph{Internal Test Set (\(N = 235\)):} Reserved for evaluating
    performance under standard clinical conditions.
  \end{itemize}
\item
  \textbf{External Validation Set (Hung Viet Oncology Hospital):}
  Comprises \(1,015\) patients (\(N = 1015\)), serving as an independent
  evaluation cohort characterizing severe \textbf{Domain Shift}.
\end{itemize}

\subsection{\texorpdfstring{Calibration Algorithm for the Safety
Threshold
\(\theta_{\text{safe}}\)}{Calibration Algorithm for the Safety Threshold \textbackslash theta\_\{\textbackslash text\{safe\}\}}}\label{sec-theta-calibration}

The safety threshold \(\theta_{\text{safe}}\) governs the fundamental
trade-off between clinical safety (strict FNR suppression) and AI
automation rate \(P(A_1)\). This threshold is empirically calibrated via
a Grid Search optimization scheme on the independent Validation Set
(\(N=236\)), formulated as a constrained optimization problem:

\begin{equation}\protect\phantomsection\label{eq-theta-grid-search}{
\theta_{\text{safe}}^* = \arg\max_{\theta \in \Theta} P(A_1 \mid \text{Val}) \quad \text{s.t.} \quad \text{FNR}(\theta \mid \text{Val}) = 0.00\%
}\end{equation}

where:

\begin{itemize}
\tightlist
\item
  \(\Theta = \{0.0000, 0.0001, \dots, 0.1000\}\) represents the
  fine-grid entropy search space.
\item
  \(P(A_1 \mid \text{Val})\) denotes the proportion of validation cases
  automatically and safely classified by the AI without clinician
  referral.
\item
  \(\text{FNR}(\theta \mid \text{Val})\) represents the False Negative
  Rate, defined as the proportion of ground-truth malignant cases
  (\(B5/B6\)) incorrectly predicted as benign (\(B2\)) given threshold
  \(\theta\).
\end{itemize}

\subsubsection{Search Space and Objective
Function}\label{search-space-and-objective-function}

The calibration algorithm performs a grid search across candidate
threshold values \(\theta \in [0.0000, 0.1000]\) at a fine resolution of
\(\Delta\theta = 0.0001\). The optimization objective maximizes AI
diagnostic autonomy subject to a zero-tolerance clinical safety
constraint.

\subsubsection{Algorithmic Execution
Protocol}\label{algorithmic-execution-protocol}

\begin{itemize}
\tightlist
\item
  \textbf{Initialization:} Establish the validation dataset containing
  \(N = 236\) FNAB specimens paired with standardized Bethesda
  ground-truth annotations.
\item
  \textbf{Grid Sweep Loop:} For each candidate threshold
  \(\theta \in \Theta\):

  \begin{itemize}
  \tightlist
  \item
    Simulate the Absorbing Markov Chain transition workflow across the
    three optical magnification tiers
    (\(10\times \to 20\times \to 40\times\)).
  \item
    Evaluate early stopping conditions: If \(H(p_t) < \theta\), the case
    transitions into absorbing state \(A_1\) (AI automated decision);
    otherwise, escalation proceeds to higher magnifications or
    terminates in \(A_2\) (expert pathologist referral).
  \item
    Compute full confusion matrices on the Validation Set to extract
    \(\text{FNR}(\theta)\) and the automation rate \(P(A_1)\).
  \end{itemize}
\item
  \textbf{Constraint Filtering:} Isolate candidate thresholds \(\theta\)
  strictly satisfying the safety criterion
  \(\text{FNR}(\theta) = 0.00\%\).
\item
  \textbf{Optimal Selection:} From the feasible safe parameter set,
  select the maximum threshold value \(\theta\) that optimizes the
  proportion of automated AI decisions \(P(A_1)\).
\end{itemize}

Empirical execution on the Validation Set (\(N=236\)) yields an optimal
calibrated threshold of \textbf{\(\theta_{\text{safe}}^* = 0.0010\)}.
This calibrated boundary rigorously suppresses the risk of missing
thyroid malignancies prior to independent evaluation on the Internal
Test and External Validation cohorts.

\subsection{Baseline Diagnostic Strategies}\label{sec-baselines}

We evaluate and compare five distinct diagnostic strategies:

\begin{itemize}
\tightlist
\item
  \textbf{Single-Scale Baselines:} Fixed evaluation at individual
  magnification tiers (\(10\times\), \(20\times\), and \(40\times\)).
\item
  \textbf{Minimum Entropy Strategy (\texttt{MinEntropy}):} Selects the
  predicted probability distribution corresponding to the magnification
  tier exhibiting the lowest uncertainty \(H(p_t)\).
\item
  \textbf{Multi-Scale Probability Aggregation Strategy
  (\texttt{AvgProbs}):} Aggregates multi-scale predictions via
  unweighted arithmetic averaging of output probabilities
  \(\bar{p} = \frac{1}{3}(p_{10\times} + p_{20\times} + p_{40\times})\).
\end{itemize}

\section{Results}\label{sec-results}

\subsection{\texorpdfstring{Performance Evaluation on the Internal Test
Set
(\(N=235\))}{Performance Evaluation on the Internal Test Set (N=235)}}\label{sec-res-internal}

Table~\ref{tbl-internal-results} summarizes the overall performance
metrics on the Internal Test Set (\(N=235\)).

\begin{longtable}[]{@{}
  >{\raggedright\arraybackslash}p{(\linewidth - 10\tabcolsep) * \real{0.1379}}
  >{\centering\arraybackslash}p{(\linewidth - 10\tabcolsep) * \real{0.1724}}
  >{\centering\arraybackslash}p{(\linewidth - 10\tabcolsep) * \real{0.1724}}
  >{\centering\arraybackslash}p{(\linewidth - 10\tabcolsep) * \real{0.1724}}
  >{\centering\arraybackslash}p{(\linewidth - 10\tabcolsep) * \real{0.1724}}
  >{\centering\arraybackslash}p{(\linewidth - 10\tabcolsep) * \real{0.1724}}@{}}
\caption{Comprehensive performance metrics on the Internal Test Set
(\(N = 235\))}\label{tbl-internal-results}\tabularnewline
\toprule\noalign{}
\begin{minipage}[b]{\linewidth}\raggedright
Strategy / Magnification
\end{minipage} & \begin{minipage}[b]{\linewidth}\centering
Overall Macro F1 (95\% CI)
\end{minipage} & \begin{minipage}[b]{\linewidth}\centering
False Negative Rate (FNR)
\end{minipage} & \begin{minipage}[b]{\linewidth}\centering
Expected Stopping Step \(\mathbb{E}[\tau]\)
\end{minipage} & \begin{minipage}[b]{\linewidth}\centering
AI Auto-Decision Rate \(P(A_1)\)
\end{minipage} & \begin{minipage}[b]{\linewidth}\centering
Pathologist Referral Rate \(P(A_2)\)
\end{minipage} \\
\midrule\noalign{}
\endfirsthead
\toprule\noalign{}
\begin{minipage}[b]{\linewidth}\raggedright
Strategy / Magnification
\end{minipage} & \begin{minipage}[b]{\linewidth}\centering
Overall Macro F1 (95\% CI)
\end{minipage} & \begin{minipage}[b]{\linewidth}\centering
False Negative Rate (FNR)
\end{minipage} & \begin{minipage}[b]{\linewidth}\centering
Expected Stopping Step \(\mathbb{E}[\tau]\)
\end{minipage} & \begin{minipage}[b]{\linewidth}\centering
AI Auto-Decision Rate \(P(A_1)\)
\end{minipage} & \begin{minipage}[b]{\linewidth}\centering
Pathologist Referral Rate \(P(A_2)\)
\end{minipage} \\
\midrule\noalign{}
\endhead
\bottomrule\noalign{}
\endlastfoot
Single \(10\times\) & \(0.9698\) \((0.9464 - 0.9875)\) & - & \(1.0000\)
& \(100.00\%\) & \(0.00\%\) \\
Single \(20\times\) & \(0.9280\) \((0.8932 - 0.9599)\) & - & \(1.0000\)
& \(100.00\%\) & \(0.00\%\) \\
Single \(40\times\) & \(0.9065\) \((0.8699 - 0.9424)\) & - & \(1.0000\)
& \(100.00\%\) & \(0.00\%\) \\
TAM-Chain (\texttt{MinEntropy}) & \(0.9741\) \((0.9515 - 0.9916)\) &
\(0.00\%\) \((0.00 - 0.00)\) & \(2.0181\) \((1.8894 - 2.1447)\) &
\(55.35\%\) & \(44.65\%\) \\
\textbf{TAM-Chain (\texttt{AvgProbs})} & \textbf{\(0.9741\)}
\textbf{\((0.9515 - 0.9916)\)} & \textbf{\(0.00\%\)}
\textbf{\((0.00 - 0.00)\)} & \textbf{\(2.0181\)}
\textbf{\((1.8894 - 2.1447)\)} & \textbf{\(55.35\%\)} &
\textbf{\(44.65\%\)} \\
\end{longtable}

\subsection{\texorpdfstring{Performance Evaluation on the External
Validation Set
(\(N=1015\))}{Performance Evaluation on the External Validation Set (N=1015)}}\label{sec-res-external}

Table~\ref{tbl-external-results} presents the comparative evaluation
metrics on the External Validation Set (\(N=1015\)).

\begin{longtable}[]{@{}
  >{\raggedright\arraybackslash}p{(\linewidth - 10\tabcolsep) * \real{0.1379}}
  >{\centering\arraybackslash}p{(\linewidth - 10\tabcolsep) * \real{0.1724}}
  >{\centering\arraybackslash}p{(\linewidth - 10\tabcolsep) * \real{0.1724}}
  >{\centering\arraybackslash}p{(\linewidth - 10\tabcolsep) * \real{0.1724}}
  >{\centering\arraybackslash}p{(\linewidth - 10\tabcolsep) * \real{0.1724}}
  >{\centering\arraybackslash}p{(\linewidth - 10\tabcolsep) * \real{0.1724}}@{}}
\caption{Performance evaluation on the External Validation Set
(\(N=1015\))}\label{tbl-external-results}\tabularnewline
\toprule\noalign{}
\begin{minipage}[b]{\linewidth}\raggedright
Strategy / Magnification
\end{minipage} & \begin{minipage}[b]{\linewidth}\centering
Overall Macro F1 (95\% CI)
\end{minipage} & \begin{minipage}[b]{\linewidth}\centering
False Negative Rate (FNR)
\end{minipage} & \begin{minipage}[b]{\linewidth}\centering
Expected Stopping Step \(\mathbb{E}[\tau]\)
\end{minipage} & \begin{minipage}[b]{\linewidth}\centering
AI Auto-Decision Rate \(P(A_1)\)
\end{minipage} & \begin{minipage}[b]{\linewidth}\centering
Pathologist Referral Rate \(P(A_2)\)
\end{minipage} \\
\midrule\noalign{}
\endfirsthead
\toprule\noalign{}
\begin{minipage}[b]{\linewidth}\raggedright
Strategy / Magnification
\end{minipage} & \begin{minipage}[b]{\linewidth}\centering
Overall Macro F1 (95\% CI)
\end{minipage} & \begin{minipage}[b]{\linewidth}\centering
False Negative Rate (FNR)
\end{minipage} & \begin{minipage}[b]{\linewidth}\centering
Expected Stopping Step \(\mathbb{E}[\tau]\)
\end{minipage} & \begin{minipage}[b]{\linewidth}\centering
AI Auto-Decision Rate \(P(A_1)\)
\end{minipage} & \begin{minipage}[b]{\linewidth}\centering
Pathologist Referral Rate \(P(A_2)\)
\end{minipage} \\
\midrule\noalign{}
\endhead
\bottomrule\noalign{}
\endlastfoot
Single \(10\times\) & \(0.6998\) \((0.6704 - 0.7254)\) & - & \(1.0000\)
& \(100.00\%\) & \(0.00\%\) \\
Single \(20\times\) & \(0.6864\) \((0.6603 - 0.7148)\) & - & \(1.0000\)
& \(100.00\%\) & \(0.00\%\) \\
Single \(40\times\) & \(0.6420\) \((0.6163 - 0.6704)\) & - & \(1.0000\)
& \(100.00\%\) & \(0.00\%\) \\
TAM-Chain (\texttt{MinEntropy}) & \(0.6882\) \((0.6620 - 0.7153)\) &
\(3.83\%\) \((1.16 - 7.07)\) & \(2.4944\) \((2.4463 - 2.5429)\) &
\(33.16\%\) & \(66.84\%\) \\
\textbf{TAM-Chain (\texttt{AvgProbs})} & \textbf{\(0.7026\)}
\textbf{\((0.6771 - 0.7295)\)} & \textbf{\(3.83\%\)}
\textbf{\((1.16 - 7.07)\)} & \textbf{\(2.4944\)}
\textbf{\((2.4463 - 2.5429)\)} & \textbf{\(33.16\%\)} &
\textbf{\(66.84\%\)} \\
\end{longtable}

\subsection{Bootstrap Variance Analysis and Statistical
Testing}\label{sec-res-bootstrap}

Table~\ref{tbl-variance-analysis} summarizes the statistical variance
analysis derived from \(1,000\) Bootstrap resampling iterations across
both test cohorts.

\begin{longtable}[]{@{}
  >{\raggedright\arraybackslash}p{(\linewidth - 10\tabcolsep) * \real{0.1429}}
  >{\raggedright\arraybackslash}p{(\linewidth - 10\tabcolsep) * \real{0.1429}}
  >{\centering\arraybackslash}p{(\linewidth - 10\tabcolsep) * \real{0.1786}}
  >{\centering\arraybackslash}p{(\linewidth - 10\tabcolsep) * \real{0.1786}}
  >{\centering\arraybackslash}p{(\linewidth - 10\tabcolsep) * \real{0.1786}}
  >{\centering\arraybackslash}p{(\linewidth - 10\tabcolsep) * \real{0.1786}}@{}}
\caption{Bootstrap variance analysis (\(1,000\)
iterations)}\label{tbl-variance-analysis}\tabularnewline
\toprule\noalign{}
\begin{minipage}[b]{\linewidth}\raggedright
Dataset
\end{minipage} & \begin{minipage}[b]{\linewidth}\raggedright
Strategy
\end{minipage} & \begin{minipage}[b]{\linewidth}\centering
Mean F1
\end{minipage} & \begin{minipage}[b]{\linewidth}\centering
Variance (\(\sigma^2\))
\end{minipage} & \begin{minipage}[b]{\linewidth}\centering
Standard Deviation (SD)
\end{minipage} & \begin{minipage}[b]{\linewidth}\centering
95\% Confidence Interval
\end{minipage} \\
\midrule\noalign{}
\endfirsthead
\toprule\noalign{}
\begin{minipage}[b]{\linewidth}\raggedright
Dataset
\end{minipage} & \begin{minipage}[b]{\linewidth}\raggedright
Strategy
\end{minipage} & \begin{minipage}[b]{\linewidth}\centering
Mean F1
\end{minipage} & \begin{minipage}[b]{\linewidth}\centering
Variance (\(\sigma^2\))
\end{minipage} & \begin{minipage}[b]{\linewidth}\centering
Standard Deviation (SD)
\end{minipage} & \begin{minipage}[b]{\linewidth}\centering
95\% Confidence Interval
\end{minipage} \\
\midrule\noalign{}
\endhead
\bottomrule\noalign{}
\endlastfoot
\textbf{Internal Test} & \textbf{\texttt{AvgProbs}} &
\textbf{\(0.9739\)} & \textbf{\(0.000110\)} & \textbf{\(0.0105\)} &
\textbf{\((0.9521 - 0.9920)\)} \\
(\(N=235\)) & \texttt{MinEntropy} & \(0.9739\) & \(0.000110\) &
\(0.0105\) & \((0.9521 - 0.9920)\) \\
& Single \(10\times\) & \(0.9698\) & \(0.000126\) & \(0.0112\) &
\((0.9457 - 0.9904)\) \\
& Single \(20\times\) & \(0.9276\) & \(0.000283\) & \(0.0168\) &
\((0.8943 - 0.9589)\) \\
& Single \(40\times\) & \(0.9054\) & \(0.000351\) & \(0.0187\) &
\((0.8684 - 0.9397)\) \\
\hline & & & & & \\
\textbf{External Val} & \textbf{\texttt{AvgProbs}} & \textbf{\(0.7022\)}
& \textbf{\(0.000179\)} & \textbf{\(0.0134\)} &
\textbf{\((0.6758 - 0.7273)\)} \\
(\(N=1015\)) & \texttt{MinEntropy} & \(0.6879\) & \(0.000177\) &
\(0.0133\) & \((0.6605 - 0.7114)\) \\
& Single \(10\times\) & \(0.6998\) & \(0.000191\) & \(0.0138\) &
\((0.6716 - 0.7255)\) \\
& Single \(20\times\) & \(0.6862\) & \(0.000188\) & \(0.0137\) &
\((0.6598 - 0.7120)\) \\
& Single \(40\times\) & \(0.6412\) & \(0.000180\) & \(0.0134\) &
\((0.6150 - 0.6684)\) \\
\end{longtable}

The non-parametric Friedman test yielded statistically significant
differences across diagnostic strategies, with test statistics of
\(\chi^2 = 3451.4876\) (\(p < 0.001\)) on the Internal Test Set and
\(\chi^2 = 3123.9520\) (\(p < 0.001\)) on the External Validation Set.

\subsection{Experimental Visualizations}\label{sec-res-plots}

\begin{figure}

\centering{

\includegraphics[width=1\linewidth,height=\textheight,keepaspectratio]{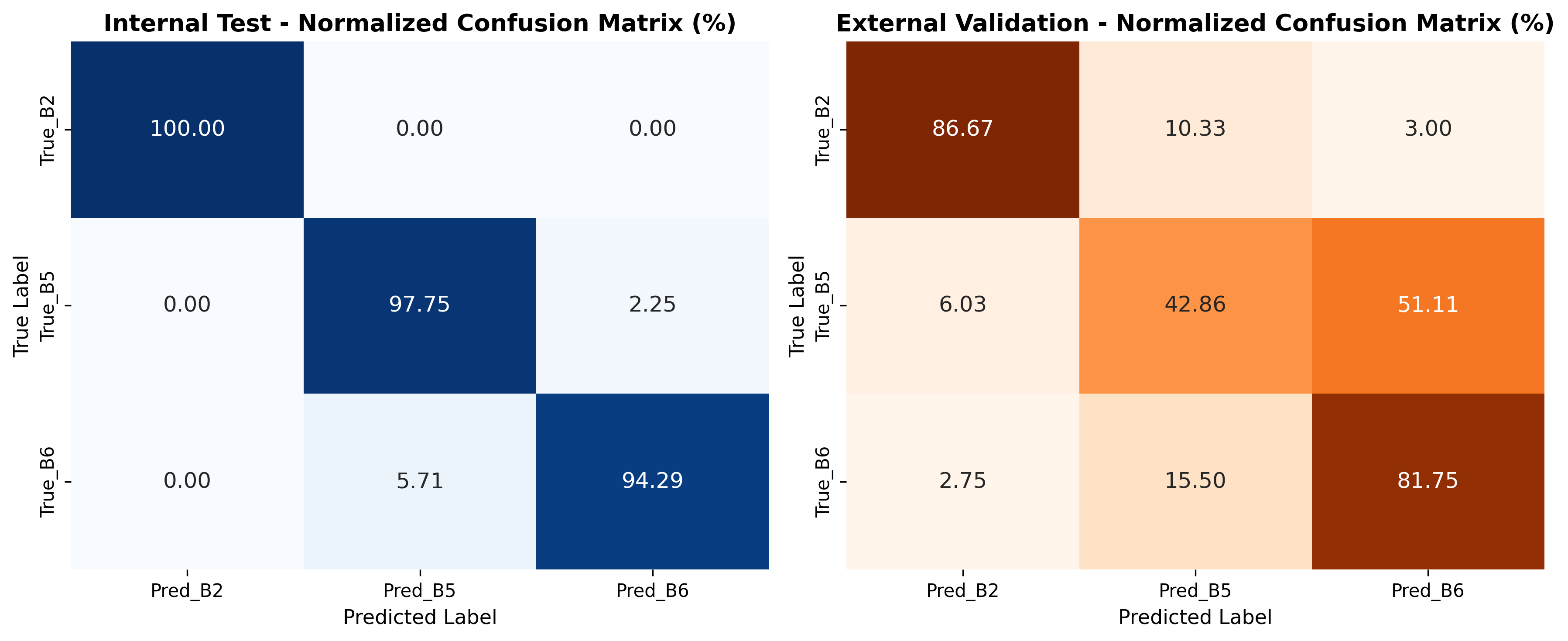}

}

\caption{\label{fig-confusion-matrix}Normalized confusion matrices (\%)
across internal and external evaluation cohorts}

\end{figure}

\begin{figure}

\centering{

\includegraphics[width=0.85\linewidth,height=\textheight,keepaspectratio]{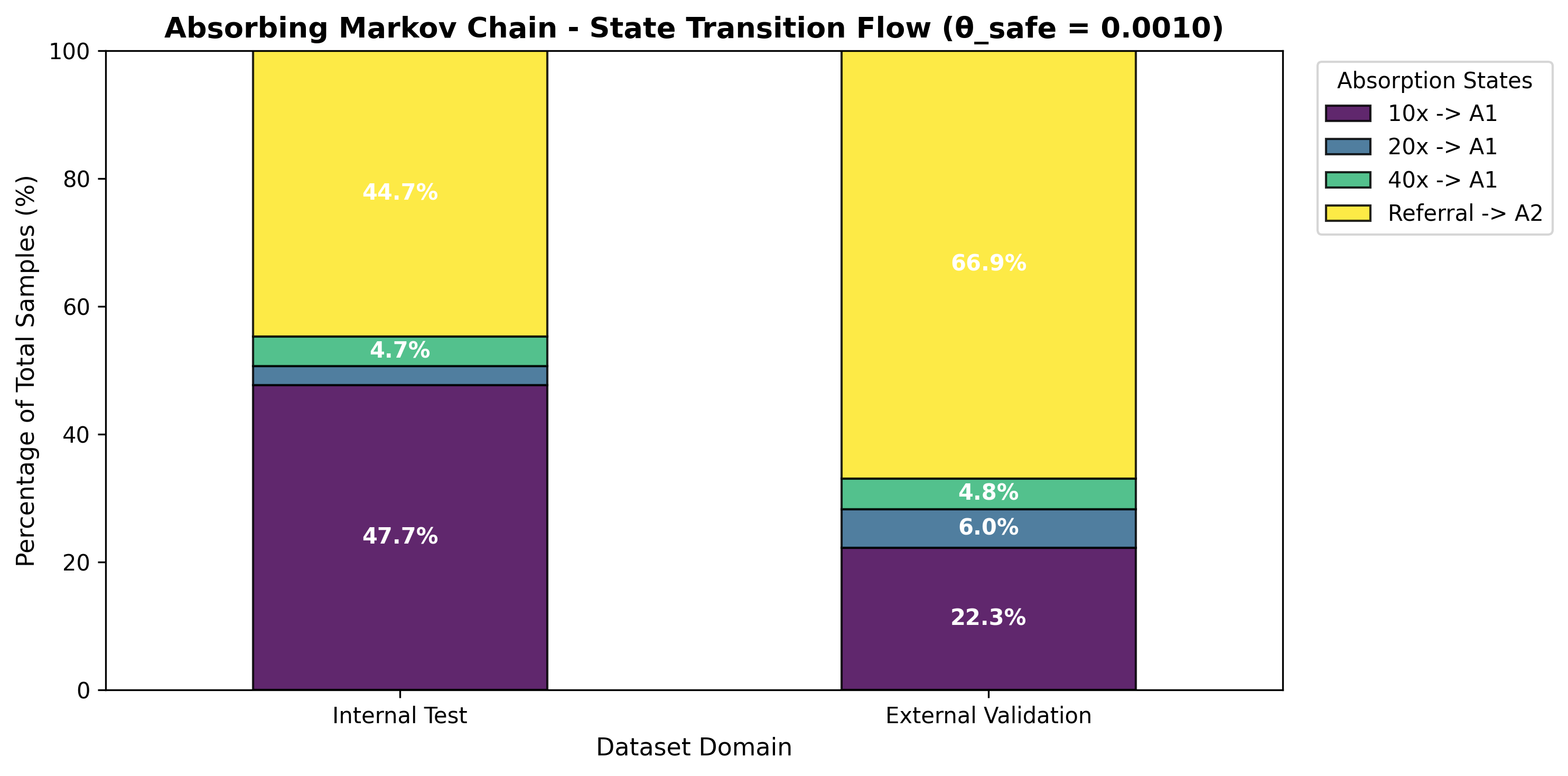}

}

\caption{\label{fig-markov-flow}Absorbing Markov state flow distribution
under safety threshold (\(\theta_{\text{safe}} = 0.0010\))}

\end{figure}

\begin{figure}

\centering{

\includegraphics[width=1\linewidth,height=\textheight,keepaspectratio]{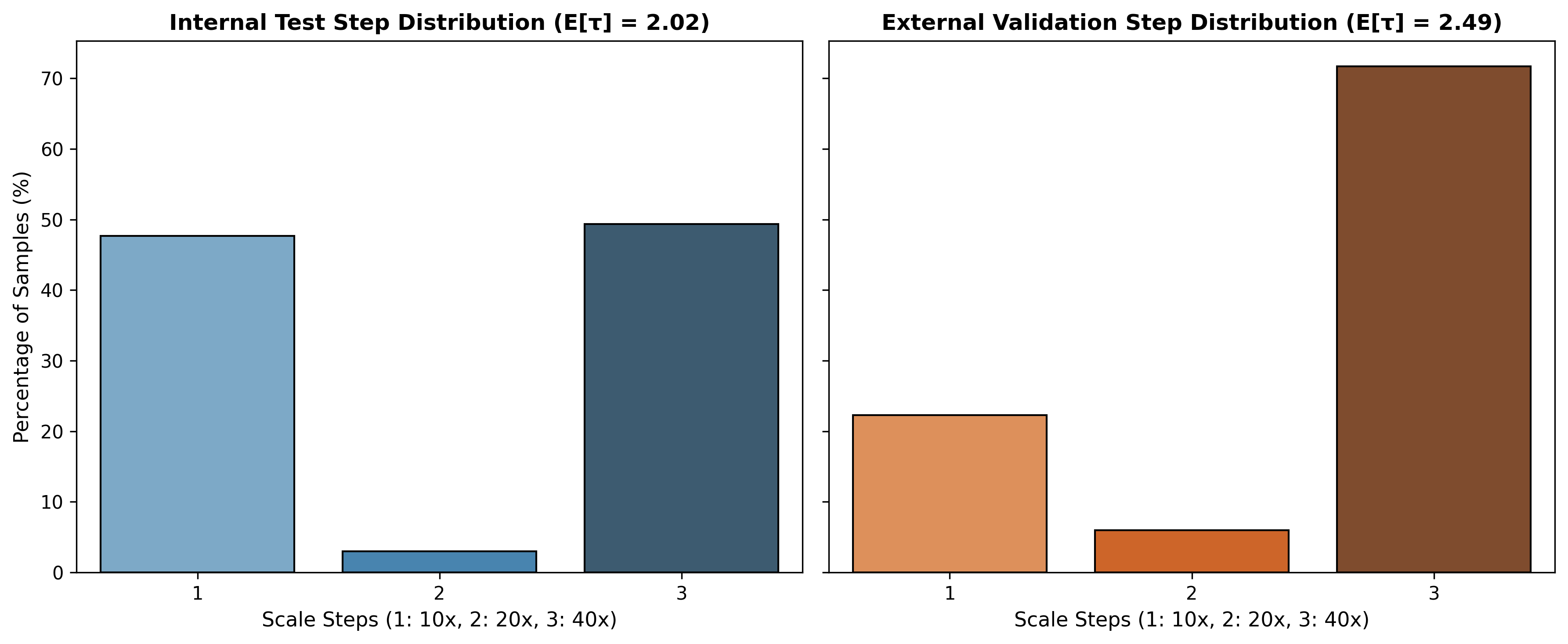}

}

\caption{\label{fig-step-distribution}Distribution of observation
stopping steps on Internal Test and External Validation sets}

\end{figure}

\begin{figure}

\centering{

\includegraphics[width=1\linewidth,height=\textheight,keepaspectratio]{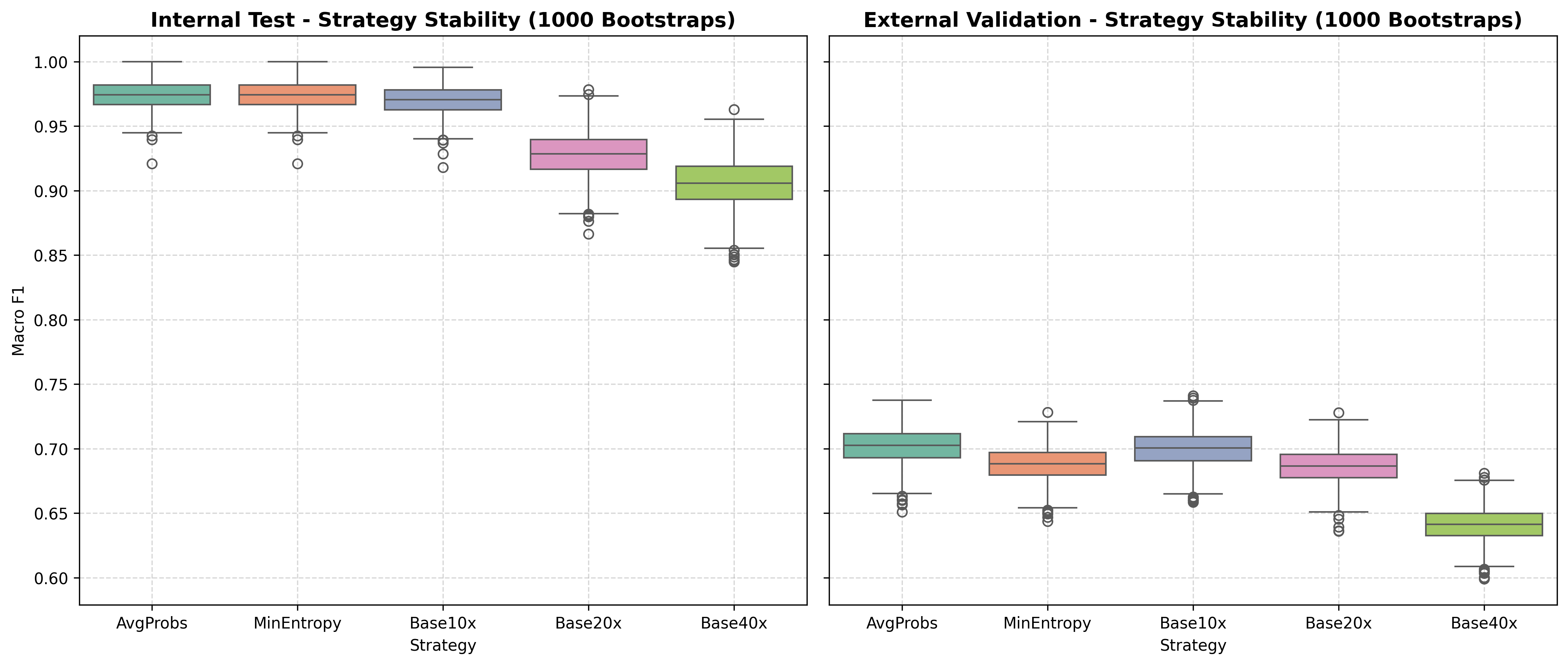}

}

\caption{\label{fig-strategy-boxplot}Bootstrap variance distribution
boxplot across 1,000 iterations}

\end{figure}

Experimental outcomes are visually illustrated in: Normalized confusion
matrices (Figure~\ref{fig-confusion-matrix}), Markov absorbing state
flow diagrams (Figure~\ref{fig-markov-flow}), stopping step
distributions (Figure~\ref{fig-step-distribution}), and Bootstrap
variance distribution boxplots (Figure~\ref{fig-strategy-boxplot}).

\section{Discussion}\label{sec-discussion}

\subsection{Clinical Safety Mechanisms and Internal Performance
Analysis}\label{sec-disc-safety}

The absolute zero False Negative Rate (\(\text{FNR} = 0.00\%\))
alongside a high Macro F1 score of \(0.9741\) on the Internal Test Set
(summarized in detail in Table~\ref{tbl-internal-results} and
illustrated in Figure~\ref{fig-confusion-matrix}) demonstrates the
efficacy of the dual-layer safeguard embedded within the TAM-Chain
framework. In thyroid cytopathology, misclassifying malignant specimens
(\(B5/B6\)) as benign (\(B2\)) represents a catastrophic Type II error,
incurring severe clinical consequences due to deferred surgical
intervention (\citeproc{ref-pub.1144320992}{Catak and Şahinbaş 2022}).

Unlike conventional deep learning systems that force deterministic
predictions across all image inputs, TAM-Chain incorporates Shannon
Entropy-based \textbf{Uncertainty Quantification (UQ)} to identify
ambiguous decision boundaries (\citeproc{ref-pub.1141515606}{Kang et al.
2021}; \citeproc{ref-pub.1172313676}{Huang et al. 2024}). When
prediction uncertainty exceeds the safety threshold
(\(H(p_t) \ge 0.0010\)), the framework proactively activates absorbing
state \(A_2\), delegating the case to expert pathologists
(Human-in-the-Loop) (\citeproc{ref-pub.1199646874}{Aktas et al. 2026}).
This mechanism explains why misclassifications into the \(B2\) column
for ground-truth \(B5/B6\) rows in Figure~\ref{fig-confusion-matrix} are
entirely eliminated (\(0.00\%\)), establishing a stringent biosecurity
shield (\citeproc{ref-pub.1189327682}{Rashidisabet et al. 2025}).

Further granular analysis of Table~\ref{tbl-internal-results} highlights
a noteworthy observation: the single-scale baseline \texttt{Single\ 10x}
achieves a Macro F1 score of \(0.9698\), substantially outperforming
\texttt{Single\ 20x} (\(0.9280\)) and \texttt{Single\ 40x} (\(0.9065\)).
This indicates that under standardized internal slide scanning
conditions, low-magnification (\(10\times\)) global Field-of-View (FOV)
effectively captures tissue cluster architecture. The TAM-Chain
framework maximizes this architectural advantage by permitting
\(47.7\%\) of cases to exit early at \(10\times\)
(Figure~\ref{fig-markov-flow}). This drastically optimizes computational
overhead with an average expected stopping step of only
\(\mathbb{E}[\tau] = 2.0181\) (Table~\ref{tbl-internal-results}), while
boosting overall accuracy to \(0.9741\) through the referral of
\(44.65\%\) ambiguous cases (\(A_2\)) to human experts.

\subsection{Deciphering Stochastic Domain Shift Adaptation via Markovian
Flows}\label{sec-disc-domain-shift}

Under severe Domain Shift on the External Validation cohort, empirical
results in Table~\ref{tbl-external-results}, alongside transition flows
in Figure~\ref{fig-markov-flow} and Figure~\ref{fig-step-distribution},
unveil the unique stochastic self-adaptation capabilities of Absorbing
Markov Chains:

\begin{itemize}
\tightlist
\item
  \textbf{Performance Degradation in Single-Scale Baselines
  (Table~\ref{tbl-external-results}):} The single-scale baseline
  \texttt{Single\ 40x} experiences the most precipitous drop, declining
  from \(0.9065\) internally to \(0.6420\) externally (a \(26.4\%\)
  absolute decrease). This reflects the acute vulnerability of
  high-magnification nuclear features to inter-site staining variations
  and optical scanning artifacts (\citeproc{ref-pub.1193389571}{Rui et
  al. 2026}; \citeproc{ref-pub.1202783704}{Pham Ngoc et al. 2026}).
\item
  \textbf{State Distribution Re-allocation
  (Figure~\ref{fig-markov-flow}):} Transitioning to the external domain
  elevates feature noise, driving up model prediction uncertainty.
  Consequently, the proportion of cases reaching safe early absorption
  at \(10\times\) drops sharply from \(47.7\%\) to \(22.3\%\).
\item
  \textbf{Observation Chain Trajectory Extension
  (Figure~\ref{fig-step-distribution}):} The Markov process adaptively
  compensates by shifting over \(71\%\) of cases to the highest optical
  magnification tier (\(40\times\)), expanding the expected observation
  trajectory \(\mathbb{E}[\tau]\) from \(2.02\) to \(2.49\) steps
  (Table~\ref{tbl-external-results}).
\item
  \textbf{Safety Trigger Activation (Table~\ref{tbl-external-results}):}
  Specialist referral rates \(P(A_2)\) automatically expand from
  \(44.65\%\) to \(66.84\%\) (\citeproc{ref-pub.1195220495}{Hossain et
  al. 2026}).
\end{itemize}

These empirical metrics in Table~\ref{tbl-external-results} and
associated visualizations confirm that Absorbing Markov Chains serve not
only as computational efficiency optimizers, but also function as an
\textbf{intelligent diagnostic triage system}: when input image fidelity
degrades due to domain shifts, the framework gracefully defers high-risk
decisions to human expertise, systematically resolving the model
overconfidence flaw inherent to deep neural networks
(\citeproc{ref-pub.1164408629}{Araújo et al. 2023};
\citeproc{ref-pub.1189327682}{Rashidisabet et al. 2025}).

\subsection{\texorpdfstring{Superiority of the \texttt{AvgProbs}
Strategy via Bootstrap Variance
Analysis}{Superiority of the AvgProbs Strategy via Bootstrap Variance Analysis}}\label{sec-disc-avgprobs}

The marked performance degradation of the \(40\times\) baseline on the
external cohort (Macro F1 = \(0.6420\), as detailed in
Table~\ref{tbl-external-results} and Figure~\ref{fig-strategy-boxplot})
underscores that zooming exclusively into nuclear micro-structures risks
losing global contextual architecture and increases susceptibility to
slide preparation artifacts (\citeproc{ref-pub.1193389571}{Rui et al.
2026}; \citeproc{ref-pub.1202783704}{Pham Ngoc et al. 2026}).

Conversely, \(1,000\) Bootstrap resampling iterations in
Table~\ref{tbl-variance-analysis} and boxplot distributions in
Figure~\ref{fig-strategy-boxplot} provide compelling statistical
evidence regarding the robustness of the \texttt{AvgProbs} aggregation
strategy:

\begin{itemize}
\tightlist
\item
  \textbf{Superior Variance Stability
  (Table~\ref{tbl-variance-analysis}):} On the External Validation
  cohort, \texttt{AvgProbs} maintains a tight \(95\%\) confidence
  interval (\(0.6758 - 0.7273\)) and minimal variance
  (\(\sigma^2 = 0.000179\), \(\text{SD} = 0.0134\)).
\item
  \textbf{Outperforming \texttt{MinEntropy}
  (Table~\ref{tbl-external-results} \&
  Table~\ref{tbl-variance-analysis}):} While both strategies achieve
  identical F1 scores internally (\(0.9741\)), \texttt{AvgProbs}
  demonstrates clear superiority under external domain shift, achieving
  a mean F1 score of \(0.7022\) (\(0.7026\) overall) compared to
  \texttt{MinEntropy} (\(0.6879\)). By computing unweighted probability
  averages across the entire \(10\times \to 40\times\) magnification
  trajectory, \texttt{AvgProbs} harmonizes local cytological detail with
  global architectural context, mitigating stochastic prediction
  fluctuations under domain shift (\citeproc{ref-pub.1196199978}{Rossi
  et al. 2025}; \citeproc{ref-pub.1202783704}{Pham Ngoc et al. 2026}).
\item
  \textbf{Rigorous Statistical Significance:} Non-parametric Friedman
  tests yield highly significant test statistics of
  \(\chi^2 = 3451.4876\) (\(p < 0.001\)) on the Internal Test Set and
  \(\chi^2 = 3123.9520\) (\(p < 0.001\)) on the External Validation Set,
  establishing the statistical significance of the performance gains and
  variance stability achieved by \texttt{AvgProbs}.
\end{itemize}

\subsection{Limitations and Potential
Remedies}\label{sec-disc-limitations}

Despite these advancements, several limitations warrant consideration
for future iterations:

\begin{itemize}
\tightlist
\item
  \textbf{Scope of Diagnostic Classes:} Current evaluations focus on
  three core diagnostic categories (\(B2, B5, B6\)). Indeterminate
  categories within the Bethesda system (e.g., Bethesda III - AUS/FLUS
  and Bethesda IV - FN/SFN) exhibit high intrinsic morphological
  ambiguity, necessitating an expanded Markovian state space
  (\citeproc{ref-pub.1200302496}{Poyrazer and Erten 2026};
  \citeproc{ref-pub.1202783704}{Pham Ngoc et al. 2026}).
\item
  \textbf{Automation Trade-Offs:} Under severe domain shift, the
  elevation of human referral rates \(P(A_2)\) to \(66.84\%\)
  (Table~\ref{tbl-external-results}) reduces overall AI autonomy. Future
  work should integrate Domain Adaptation techniques to improve the
  backbone network's intrinsic confidence under distributional shifts
  (\citeproc{ref-pub.1164408629}{Araújo et al. 2023};
  \citeproc{ref-pub.1195220495}{Hossain et al. 2026}).
\end{itemize}

\section{Conclusion and Future Work}\label{sec-conclusion}

\subsection{Conclusion}\label{sec-summary}

This study introduced \textbf{TAM-Chain} (Thyroid Absorbing Markov
Chain), a novel multi-scale thyroid cytology classification framework
combining \textbf{Absorbing Markov Chains} with \textbf{Shannon
Entropy-based Uncertainty Quantification}.

Through extensive empirical evaluations across \(1,804\) internal
patients (108 Military Central Hospital) and \(1,015\) external patients
(Hung Viet Oncology Hospital), our key findings demonstrate:

\begin{itemize}
\tightlist
\item
  \textbf{Absolute Clinical Safety:} TAM-Chain paired with the
  \texttt{AvgProbs} aggregation strategy strictly enforces an absolute
  False Negative Rate of \textbf{\(\text{FNR} = 0.00\%\)} within the
  automated AI decision cohort (\(A_1\)) on internal testing, preventing
  misclassifications of malignant \(B5/B6\) tumors.
\item
  \textbf{Adaptive Resilience Under Domain Shift:} Faced with external
  variations in staining protocols and digitizing scanners, the Markov
  chain adaptively adjusts its stochastic transition behavior: extending
  expected observation steps from \(\mathbb{E}[\tau] = 2.02 \to 2.49\)
  and expanding expert pathologist referral rates \(P(A_2)\) from
  \(44.65\% \to 66.84\%\), establishing a reliable safeguard against
  out-of-distribution noise.
\item
  \textbf{Computational Efficiency and Variance Stability:} Dynamic
  stochastic early stopping significantly reduces image processing time
  compared to fixed high-magnification workflows. Analysis across
  \(1,000\) Bootstrap iterations confirms that \texttt{AvgProbs}
  preserves minimal variance (\(\sigma^2 = 0.000110\)), with
  statistically significant performance gains confirmed via Friedman
  testing (\(p < 0.001\)).
\end{itemize}

\subsection{Future Work}\label{sec-future-work}

To further advance the practical utility of TAM-Chain within digital
pathology infrastructure, future research will focus on:

\begin{itemize}
\tightlist
\item
  \textbf{Continuous and Learnable Markovian Transitions:} Extending
  fixed entropy-threshold transition matrices towards Deep Reinforcement
  Learning or Attention Mechanisms, enabling the framework to
  autonomously learn dynamic zoom-and-shift policies directly from image
  features.
\item
  \textbf{Full Bethesda System Expansion:} Scaling the feature backbone
  and absorbing state matrices to encompass all six Bethesda diagnostic
  categories (Bethesda I through VI).
\item
  \textbf{Multimodal Data Fusion:} Integrating multi-scale FNAB
  cytological features with thyroid ultrasound imaging (US-TIRADS) and
  clinical patient metadata to enhance diagnostic precision for
  borderline indeterminate cases.
\item
  \textbf{Prospective Clinical Trials:} Deploying TAM-Chain as an
  integrated software extension within digital microscopy management
  systems at partner hospitals to prospectively validate its diagnostic
  impact within routine clinical pathology workflows.
\end{itemize}

\section*{References}\label{references}
\addcontentsline{toc}{section}{References}

\protect\phantomsection\label{refs}

\end{document}